\documentclass[lettersize,journal]{IEEEtran}
\usepackage{amsmath,amsfonts}
\usepackage{graphicx}
\usepackage{cite}
\usepackage{algorithm}
\usepackage{algpseudocode}
\usepackage{hyperref}

\begin{document}

\title{RANT: Ant-Inspired Multi-Robot Rainforest Exploration Using Particle Filter Localisation and Virtual Pheromone Coordination}

\author{
    Ameer Alhashemi, Layan Abdulhadi, Karam Abuodeh,\\
    Tala Baghdadi, Suryanarayana Datla\\[6pt]
}

\maketitle
\begin{abstract}
This paper presents RANT, an ant-inspired multi-robot exploration framework for noisy, uncertain environments. A team of differential-drive robots navigates a $10\times10$\,m terrain, collects noisy probe measurements of a hidden richness field, and builds local probabilistic maps while the supervisor maintains a global evaluation. RANT combines particle-filter localisation, a behaviour-based controller with gradient-driven hotspot exploitation, and a lightweight no-revisit coordination mechanism based on virtual pheromone blocking. We experimentally analyse how team size, localisation fidelity, and coordination influence coverage, hotspot recall, and redundancy. Results show that particle filtering is essential for reliable hotspot engagement, coordination substantially reduces overlap, and increasing team size improves coverage but yields diminishing returns due to interference.
\end{abstract}


\section{Introduction}
\IEEEPARstart{M}{ulti-robot} exploration is widely used in environmental monitoring, precision agriculture, and large-scale surveying, where several robots must cooperatively explore an area and gather informative measurements \cite{ref1}. Real deployments rely on noisy GPS, IMU, and wheel odometry, and often operate with limited communication; effective systems therefore favour decentralised, locally reactive controllers \cite{ref1,ref6}.

Swarm robotics provides simple decentralised rules, including pheromone-like
stigmergy, that improve coverage and reduce interference in multi-agent
exploration \cite{ref4,ref5,ref11}.

A second challenge is localisation: meaningful sampling requires each robot to maintain a stable pose estimate under noise and slip. Particle-filter (Monte Carlo) localisation provides this capability by propagating multiple pose hypotheses and updating them using motion and GPS/heading measurements \cite{ref6,ref7,ref8}. 

This work introduces \textbf{RANT}, an ant-inspired multi-robot exploration framework implemented in Webots. Robots fuse GPS, IMU, and odometry using a particle filter, explore using a behaviour-based controller, and switch to gradient-ascent exploitation when detecting high-richness areas. A lightweight supervisor provides noisy samples from a hidden richness field and broadcasts blocked zones whenever a hotspot is found, discouraging redundant revisits and promoting spatial diversity.

\subsection*{Related Work}
Classical exploration frameworks such as frontier-based exploration \cite{ref1} and occupancy-grid mapping \cite{ref2} have been extended to multi-robot settings, whereas swarm-based methods emphasise decentralised coordination through local interactions and bio-inspired rules \cite{ref4,ref5,ref10,ref11}. Distributed gradient climbing and coverage control \cite{ref9,ref13,ref12} demonstrate how simple robots can maintain spatial spread while moving toward informative regions.

Monte Carlo localisation is widely used outdoors to mitigate GPS drift and sensor noise \cite{ref6,ref7,ref8}, and visual–inertial methods are often used when GPS is degraded, for example under canopy cover \cite{ref14}. Our work combines these perspectives by analysing how PF localisation, ant-like exploration rules, and virtual-pheromone blocking influence multi-robot coverage.
\subsection*{Contributions}
We present (i) a complete Webots-based framework combining PF localisation,
gradient exploitation, and virtual-pheromone blocking; (ii) an evaluation of
team size, localisation fidelity, and coordination; and (iii) a quantitative
study of how these factors shape coverage, hotspot recall, and redundancy, under realistic noise models.

\section{System Overview}

\subsection{Webots Environment and Hidden Richness Field 
}
\label{sec:env_richness}

The experiments are conducted in a $10{\times}10\,\mathrm{m}$ Webots world
\cite{ref15} spanning $[x,y]\!\in\![-5,5]^2$ with mixed vegetation, rocks, and
natural occlusions. The terrain uses Webots' \texttt{UnevenTerrain} mesh to
introduce realistic slopes and height discontinuities, increasing motion and sensing uncertainty.

\begin{figure}[t]
    \centering
    \includegraphics[width=1\linewidth]{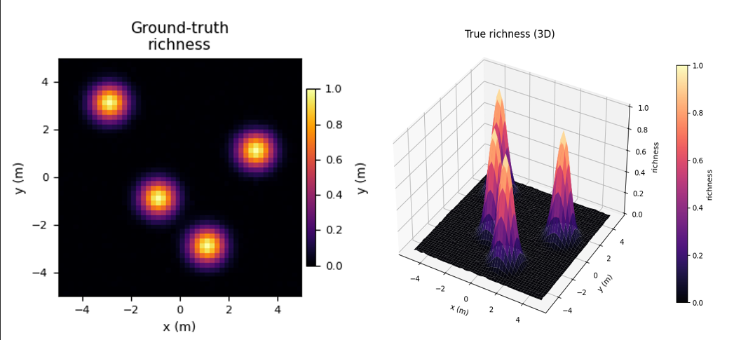}
    \caption{Supervisor visualisations: (left) ground‐truth richness $R_g$;
    (right) 3D true richness.}
    \label{fig:groundtruthplots}
\end{figure}

A hidden scalar ``richness'' field is maintained by the supervisor on a
$50{\times}50$ grid,
\begin{equation}
R_g : \{0,\ldots,49\}^2 \rightarrow [0,1],
\end{equation}
constructed as a normalised sum of four smooth Gaussian blobs with small
perturbations (Fig.~\ref{fig:groundtruthplots}). Robots never observe $R_g$
directly. During exploration, each robot requests a noisy sample at its current
estimated pose; the supervisor maps world coordinates to grid indices,
returns the corresponding scalar value, and provides a finite-difference
gradient estimate with boundary clamping. These gradient hints inform local
exploitation behaviour when robots enter BLOB mode.

\subsection{RANT Robot Platform and Sensors 
}

Each agent is a Koala-based differential-drive robot with six wheels arranged
as three linked pairs. The controller exposes wheel encoders, GPS, a
compass/IMU, and sixteen IR proximity sensors (\texttt{ds0--ds15}).
In this section we only outline the sensing suite; the full odometry and noise
models are described in Section~3.2.

\paragraph*{Odometry}
Encoder differences yield left/right wheel displacements and an incremental
heading change, following the standard differential-drive model
\cite{ref15}. Wheel accelerations are limited for stable
dynamics.

\paragraph*{GPS and Heading}
GPS and IMU measurements are perturbed with Gaussian noise to emulate realistic
outdoor uncertainty:
\begin{equation}
(x_{\text{gps}}, y_{\text{gps}}) = (x,y) + \mathcal{N}(0,0.02^2),\quad
\end{equation}
\begin{equation}
\theta_{\text{meas}} = \theta + \mathcal{N}(0,0.05^2).
\end{equation}
These noisy readings are fused later by the particle filter.

\paragraph*{Obstacle Avoidance}
The sixteen IR sensors are arranged in front-left, front-right, and lateral
arcs, providing short-range proximity information; the controller uses these
arcs for obstacle and wall avoidance as detailed in Section~\ref{sec:obstacle_handling}.

\subsection{Environmental Stress Simulation }

Three environmental modes are modelled: \texttt{clear}, \texttt{fog}, and
\texttt{rain}. The supervisor sets fog visibility, rain intensity, and slip
parameters, and stores them in \texttt{environment\_state.json}. Each robot
reads this state and adapts behaviour accordingly.

\begin{itemize}
    \item \textbf{Fog:} reduces speed, increases motion noise and sampling frequency.
    \item \textbf{Rain:} increases slip and motion noise, slightly boosts gradient gains,
    raises sampling frequency, and introduces radio dropouts.
\end{itemize}

\subsection{Supervisor, Communication, and Shared Maps 
}

The supervisor maintains the global state of the experiment and exposes a set
of shared maps over the $50{\times}50$ grid:
\begin{itemize}
    \item richness field $R_g$,
    \item pheromone map $P$ encoding no-blob-revisit behaviour,
    \item visited mask $V$,
    \item blocked mask $B$ marking confirmed hotspot exclusion zones,
    \item visit–count map $C$.
\end{itemize}
These maps form the medium through which robots indirectly influence one
another—mirroring ant-style stigmergy. Pheromone accumulation and evaporation
shape long-term coverage, while newly declared blocked regions prevent repeated
exploitation of discovered hotspots and promote spatial spread across the
swarm.

Robots exchange only minimal information with the supervisor through a
lightweight message protocol used to support sampling and global coordination:

\textbf{Robot $\rightarrow$ Supervisor}
\begin{itemize}
    \item \texttt{SAMPLE\_REQUEST}$(x,y)$ — request a noisy richness reading
          and gradient;
    \item \texttt{BULLSEYE} — announce a detected hotspot centre;
    \item \texttt{TELEMETRY} — send particle-filter pose estimates.
\end{itemize}

\textbf{Supervisor $\rightarrow$ Robot}
\begin{itemize}
    \item \texttt{SAMPLE\_RESPONSE}$(R,\nabla R)$ — return noisy values;
    \item \texttt{BLOCK}$(x,y,r)$ — publish a circular forbidden region after a hotspot is confirmed;
    \item \texttt{STOP\_ALL} — terminate the run after four hotspots are found.
\end{itemize}

Each confirmed sample at cell $(i,j)$ updates the shared maps:
\begin{equation}
P_{ij}\leftarrow P_{ij}+1,\qquad
C_{ij}\leftarrow C_{ij}+1,\qquad
V_{ij}\leftarrow 1,
\end{equation}
with pheromone evaporating as
\begin{equation}
P_{ij}(t{+}1) = 0.995\,P_{ij}(t).
\end{equation}

\begin{figure}[t]
    \centering
    \includegraphics[width=\linewidth]{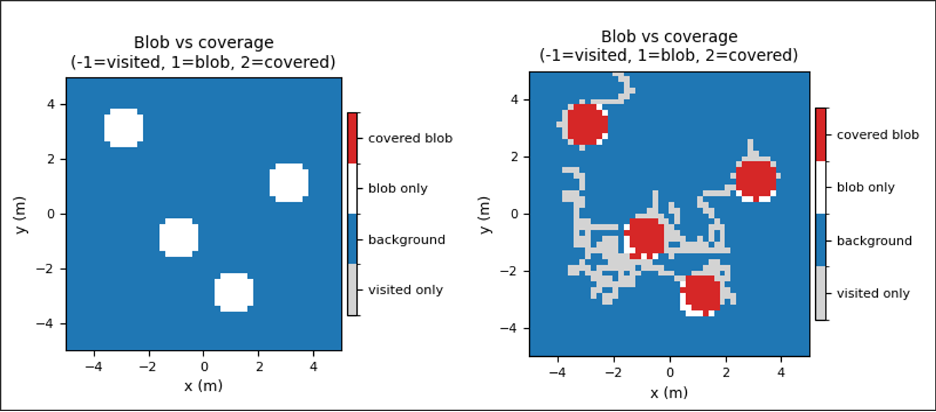}

\caption{
Supervisor blob--vs--coverage maps before and after a sample $N{=}5$ run. 
Red cells denote covered hotspot regions; white cells are hotspot cells not visited; 
grey traces show robot trajectories; blue is background. 
The internal mask encodes each grid cell as $-1$ (visited background),
$1$ (hotspot), or $2$ (covered hotspot).
}
  \label{fig:coverage_mask_plot}
\end{figure}

When a robot reports a hotspot, the supervisor inserts a blocked disc into $B$
and increments a global hotspot counter. Once four hotspots have been validated,
it broadcasts \texttt{STOP\_ALL} and constructs a blob–vs–coverage mask
(Fig.~\ref{fig:coverage_mask_plot}) summarising which hotspot cells were
visited and how trajectories intersected the ground-truth richness field.

\section{Algorithms}

This section outlines the main algorithms in RANT: construction of the hidden
richness field and local belief maps, particle-filter localisation, the
behaviour-based controller, and simple environmental adaptation. For each
component we briefly state cost, convergence behaviour, and main failure modes. 

\textit All results were generated using fixed evaluation protocols across runs. Key environment and algorithm parameters are recorded in the experiment configuration files used for evaluation.

\subsection{Richness Field, Gradients, and Local Belief Maps }

The supervisor represents the hidden richness field $R_g$ on a
$50{\times}50$ grid over the $10{\times}10\,\mathrm{m}$ arena. It is
constructed as a mixture of four Gaussian bumps,
\begin{equation}
R_g(j,i)=\sum_{k=1}^{4} 
\exp\!\left(-\frac{(i-\mu_{x,k})^2 + (j-\mu_{y,k})^2}{2\sigma^2}\right),
\end{equation}
normalised to $[0,1]$ and perturbed with small noise. Robots never see $R_g$
directly; they issue \texttt{SAMPLE\_REQUEST}$(x,y)$ messages, which the
supervisor maps to indices $(i,j)$ and answers with a noisy scalar sample and a
finite-difference gradient estimate.

World coordinates $(x,y)$ are mapped linearly to grid indices
$(i,j) \in \{0,\dots,49\}^2$ over
$[x_{\min},x_{\max}]\times[y_{\min},y_{\max}]$. Central differences with
boundary clamping approximate the gradient:
\begin{equation}
\frac{\partial R}{\partial x}\approx
\frac{R_g[j,i+1]-R_g[j,i-1]}{2\,\Delta x},
\end{equation}

\begin{equation}
\frac{\partial R}{\partial y}\approx
\frac{R_g[j+1,i]-R_g[j-1,i]}{2\,\Delta y}
\end{equation}
Cells inside supervisor-declared blocked regions return $R{=}0$ and
$\nabla R{=}0$, so confirmed hotspots become repulsive for the goal selector.

Each robot maintains a small local belief grid (40×40 cells at $\Delta = 0.1\,\mathrm{m}$
resolution) storing private, running means of sampled richness for offline inspection. For a sample with
measured richness $R_{\mathrm{meas}}$ at $(x,y)$ inside the window, the
corresponding cell $(i,j)$ stores an integer visit count $c_{ij}$ and a running
mean $M_{ij}$,
\[
c_{ij} \leftarrow c_{ij}+1,\qquad
M_{ij} \leftarrow \frac{c_{ij}\,M_{ij} + R_{\mathrm{meas}}}{c_{ij}+1}.
\]
\textit{Cost, convergence, failure modes.} $R_g$ is precomputed once in
$O(G_xG_y)$ and each query is $O(1)$. Running means converge under unbiased
noise, but boundary clamping, flat gradients, and localisation drift can blur
reconstructed blobs.

\subsection{Particle-Filter Localisation}

Each robot runs a particle filter (PF) with $N=150$ particles to estimate its
pose $(x,y,\theta)$. On the first valid GPS fix, particles are initialised in a
Gaussian window around the GPS position and IMU heading with uniform weights.

At each control cycle the PF executes a standard predict–update–resample
sequence. The motion model uses odometry-derived forward motion $d$ and
rotation $\Delta\theta$:
\[
\theta'_i = \theta_i + \Delta\theta + \mathcal{N}(0,\sigma_\theta^2),\]
\[x'_i = x_i + d\sin\theta'_i + \mathcal{N}(0,\sigma_d^2),\]
\[y'_i = y_i + d\cos\theta'_i + \mathcal{N}(0,\sigma_d^2),
\]
with $\sigma_d$ and $\sigma_\theta$ proportional to $|d|$ and $|\Delta\theta|$
to capture slip and unmodelled dynamics.

The measurement update uses the noisy GPS and heading measurements defined in
Section~II B. For each particle $(x_i,y_i,\theta_i)$ we evaluate a positional
likelihood with standard deviation $\sigma_{\mathrm{gps}}$ and, when heading is
available, a heading likelihood with standard deviation $\sigma_\theta$. Final
weights $w_i = w_i^{\mathrm{pos}} w_i^{\theta}$ are normalised so that
$\sum_i w_i = 1$, and the pose estimate is taken as the weighted mean
\begin{equation}
(x_{\mathrm{est}},y_{\mathrm{est}},\theta_{\mathrm{est}}) =
\sum_i w_i (x_i,y_i,\theta_i).
\end{equation}

To avoid particle impoverishment we monitor the effective particle count
\begin{equation}
N_{\mathrm{eff}}=\frac{1}{\sum_i w_i^2},
\end{equation}
and trigger systematic resampling whenever $N_{\mathrm{eff}} < \alpha N$ with
$\alpha = 0.95$. 

For later evaluation the controller logs the instantaneous localisation error
\begin{equation}
e(t)=\sqrt{(x_{\mathrm{true}}-x_{\mathrm{est}})^2+
(y_{\mathrm{true}}-y_{\mathrm{est}})^2},
\label{eq:loc_error}
\end{equation}
from which MAE, RMSE, and maximum error are computed in
Section~\ref{sec:pf_results}.

\textit{Cost, convergence, failure modes.} Each update is $O(N)$. With intermittent
GPS the cloud quickly settles and then shows expand–collapse cycles; if GPS
degrades or resampling is too infrequent the cloud diffuses and drift grows to
metre scale.

\subsection{Behaviour-Based Controller and Obstacle Handling }
\label{sec:obstacle_handling}
Each robot runs a three-mode finite-state controller with an emergency
\emph{unstuck} routine. Commands are issued as forward speed
$v_{\mathrm{cmd}}$ and angular velocity $w_{\mathrm{cmd}}$, then converted to
wheel speeds with acceleration limits.

\textbf{EXPLORE} is a biased random walk:
every $\sim 0.8$\,s a new exploratory term $w_{\mathrm{osc}}$ is sampled from a
zero-mean Gaussian while forward speed stays near \texttt{BASE\_FWD}. Steering
combines (i) the random term, (ii) obstacle and corner avoidance from the 16 IR
sensors, (iii) soft wall recentering, (iv) a small radial-band dispersion bias,
and (v) repulsion from locally stored blob centres and supervisor-declared
blocked discs, which act as soft virtual obstacles.

\textbf{BLOB} mode performs local search around promising samples. EXPLORE
switches to BLOB once the best observed richness exceeds
\texttt{BLOB\_ENTER\_VAL}. If a best point $(x_b,y_b)$ is known, steering tracks
the bearing $\theta_{\mathrm{target}}=\mathrm{atan2}(y_b-y,x_b-x)$ with a
bounded proportional heading correction; otherwise the supervisor’s gradient
direction is used. Forward speed decreases with heading error, and an
edge-escape term prevents sliding along blob boundaries. BLOB terminates when
values fall below \texttt{BLOB\_EXIT\_VAL} for
\texttt{BLOB\_EXIT\_COUNT} consecutive samples, after which the blob centre is
added to a private forbidden list.

\textbf{RECOVER} is entered after leaving a blob or declaring a bullseye. The
robot drives mostly straight at \texttt{RECOVER\_SPEED} using only obstacle,
wall, and no-revisit corrections, until it has moved at least
\texttt{RECOVER\_MIN\_DIST} or a timeout occurs. This prevents tight circling
near discovered hotspots. When global blocking is enabled, the newly confirmed blob centre is added to
the supervisor’s \texttt{blocked} mask, so RECOVER automatically treats the blob
region as a repulsive zone; with blocking disabled, robots simply move away
without any long-term repulsion.

An \textbf{unstuck} manoeuvre is triggered if the robot’s pose has not changed
more than \texttt{STUCK\_POS\_EPS} over \texttt{STUCK\_TIME}. It executes a
short reverse-and-turn followed by a forward turn, temporarily overriding all
other steering and clearing rare deadlocks.

Obstacle and wall handling use the 16 IR sensors grouped into front-right,
front-left, and lateral arcs. If either front arc exceeds a threshold, a
turning command $w_{\mathrm{obs}}$ is generated away from the stronger side and
forward speed is reduced as readings approach a hard-stop level; a
“corner’’ condition (both arcs high) briefly boosts turn rate. Side sensors
provide gentle lateral repulsion that keeps the robot from grazing walls.

\textit{Cost, convergence, failure modes.}
Closed-loop behaviour typically yields wide-area coverage with repeated
convergence toward unexplored hotspots until the supervisor triggers
\texttt{STOP\_ALL}. Transient jitter from noisy side sensors or tight corners
is normally resolved by RECOVER or the unstuck routine within a few seconds.
Tuning of virtual blocks is critical: overly large \texttt{BLOCK\_RADIUS\_M}
can render regions effectively inaccessible and depress coverage, whereas
very small radii or weak pheromone penalties produce behaviour almost
indistinguishable from the uncoordinated baseline, offering little reduction
in redundancy.

\subsection{Environmental Adaptation }

Robots adapt controller gains to fog and rain by reducing speed, increasing
motion noise, adjusting sampling frequency, and injecting radio dropouts under
heavy rain. In clear conditions they use the nominal maximum speed and a
0.5\,s sampling period; fog reduces speed by roughly 20--25\% and increases
sample interval, while rain halves the maximum speed, slightly increases
sampling frequency, and introduces frequent communication drops. These changes
primarily raise traversal time rather than fundamentally altering the controller
logic.

\textit{Cost, convergence, failure modes.}
Adaptation is $O(1)$ per step and mainly affects traversal time: heavy fog or
extreme slip slow coverage and delay recovery from wall contacts, but in our
runs did not produce catastrophic failures.

\section{Experimental Evaluation}

We evaluate how team size, localisation fidelity, and pheromone-based
coordination affect exploration performance and mapping accuracy. All runs use
the same $10\times10\,\mathrm{m}$ Webots arena, richness field, and noise
models.

\subsection{Experiment 1: Effect of Team Size on Coverage and Mapping Time }

\textbf{Objective.}
Quantify how increasing team size affects spatial coverage, hotspot detection,
and redundancy.

\textbf{Setup.}
We compare teams of size $N \in \{1,3,5\}$. All robots use
particle-filter localisation, the EXPLORE/BLOB/RECOVER controller, fixed,
slightly offset starting poses in a common spawn region, and the same richness
field with four Gaussian blobs.

\textbf{Hypotheses.}
\begin{itemize}
    \item \textbf{H1a (coverage and recall):} Increasing $N$ increases world
    coverage and blob recall (fraction of blob cells visited).
    \item \textbf{H1b (redundancy):} Pheromone-like no-revisit and blocked-zone
    coordination prevent redundancy from growing linearly with $N$, so
    multi-robot runs remain more sample-efficient than a naïve
    “more robots = more overlap’’ baseline.
\end{itemize}

\textbf{Metrics.}
For each $N$ we measure:
\begin{itemize}
    \item world coverage (fraction of grid cells ever visited) and blob recall;
    \item precision and F1 on coverage, treating blob cells as positives;
    \item blob detection rate (blobs with at least one visited cell);
    \item redundancy,
    \begin{equation}
      \text{redundancy} =
      1 - \frac{\text{unique visited cells}}{\text{total samples}}.
      \label{redundancy}
    \end{equation}
\end{itemize}
The four blobs occupy $180$ of $2500$ cells ($\approx 7.2\%$), so a purely
uniform sampler would achieve only $\sim 7\%$ precision.

\subsection{Experiment 2: Impact of Localisation Fidelity }

\textbf{Objective.}
Assess how localisation fidelity affects (i) the internal consistency of the
particle filter and (ii) the ability of robots to detect and exploit hotspots.

\textbf{Setup.}
We reuse the arena, richness field, and team size from
Experiment~1 and compare:
\begin{enumerate}
    \item \textbf{PF-enabled (PF--on)} 
    \item \textbf{PF-disabled (PF--off)} 
\end{enumerate}

\textbf{Hypotheses.}
\begin{itemize}
    \item \textbf{H2a (hotspot reactivity):} PF-enabled robots will enter BLOB
    mode and dwell inside hotspots; PF-disabled robots will traverse hotspots
    without reacting.
    \item \textbf{H2b (sampling and coverage):} PF-enabled runs will generate
    many valid \texttt{SAMPLE\_REQUEST}s and dense sampling around hotspots,
    whereas PF-disabled runs will produce no samples and no coverage in
    supervisor maps.
\end{itemize}

\textbf{Metrics and assessment tools.}
For each condition we analyse:
\begin{itemize}
    \item console traces of \texttt{control} and \texttt{SAMPLE\_AT}
    (finite values vs.\ \texttt{(nan, nan)});
    \item mode transitions (EXPLORE vs.\ BLOB/RECOVER);
    \item counts and spatial distribution of \texttt{SAMPLE\_REQUEST}s
    and visited cells;
    \item instantaneous localisation error (Eq.~\eqref{eq:loc_error}), from
    which MAE, RMSE, and Max error are derived;
    \item supervisor particle-cloud visualisations: stable filters show
    expand--collapse cycles as GPS updates arrive, whereas unstable filters
    diffuse and drift away from the true pose.
\end{itemize}

\subsection{Experiment 3: Influence of Pheromone Blocking and Coordination }

\textbf{Objective.}
Evaluate whether virtual pheromone blocking
(\texttt{ENABLE\_GLOBAL\_BLOCKING}) reduces redundant sampling and improves
spatial spread for a fixed team of five robots.

\textbf{Setup.}
We fix the environment, supervisor configuration, and team size to match the
final implementation. All robots use particle-filter localisation and the
EXPLORE/BLOB/RECOVER controller. We set $N=5$ to evaluate coordination effects
under realistic multi-robot interference. Two conditions are compared:
\begin{enumerate}
    \item \textbf{Coordination ON:} global blocked zones and pheromone penalties
    are applied after each \texttt{BULLSEYE}.
    \item \textbf{Coordination OFF:} no global blocking and no long-term
    repulsion; \texttt{BLOCK} messages are ignored.
\end{enumerate}

\textbf{Hypotheses.}
\begin{itemize}
    \item \textbf{H3a (redundancy):} Blocking and no-revisit should reduce the
    fraction of samples taken on already-visited cells.
    \item \textbf{H3b (hotspot discovery):} Coordination should maintain or
    improve blob recall and the number of distinct blobs discovered.
\end{itemize}

\textbf{Metrics.}
For coordination ON and OFF we compute:
\begin{itemize}
    \item world coverage and blob recall;
    \item precision and F1 on coverage (blob cells as positives);
    \item blob detection rate (blobs with at least one visited cell);
    \item redundancy Eq.~\eqref{redundancy}
    \item visit-entropy and, for coordination ON, the fraction of cells marked
    as blocked in the global \texttt{blocked} mask.
\end{itemize}

\section{Results and Discussion }

This section presents the outcomes of the three experiments. Across all runs,
we focus on coverage, hotspot recall, localisation error, and sampling redundancy.

\subsection{Effect of Team Size}

\begin{figure}[t]
    \centering
    \includegraphics[width=1\linewidth]{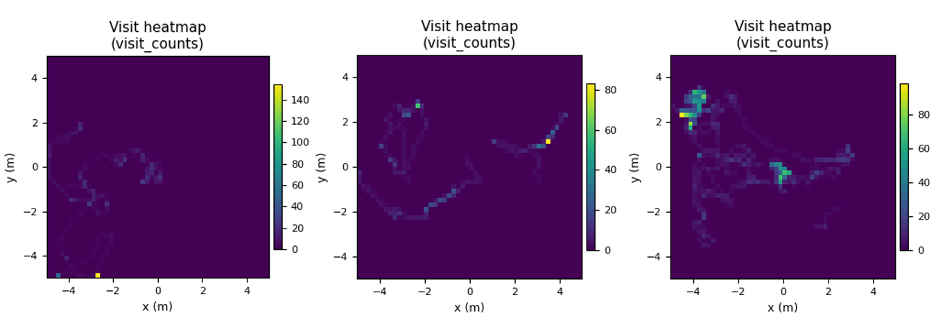}
    \caption{
Visit–count heatmaps for team sizes $N=\{1,3,5\}$ (left to right).
Brighter regions indicate cells visited more frequently.
With more robots the explored area grows and hotspots are detected earlier,
but redundant sampling becomes more localised around blob regions, reflecting
increased interference at higher team sizes.
    }
    \label{fig:visit_heatmap_team_size}
\end{figure}

\begin{figure}[t]
    \centering
    \includegraphics[width=\linewidth]{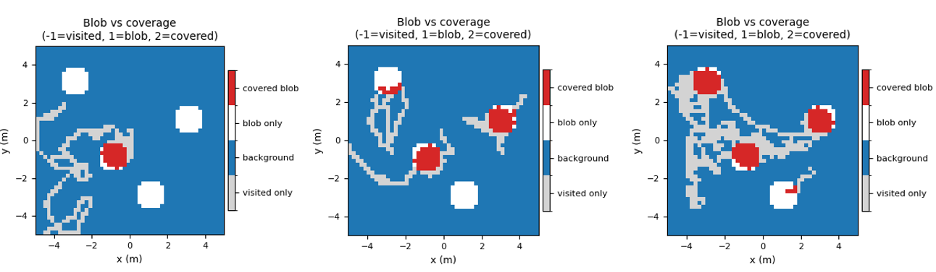}
    \caption{
Blob–vs–coverage maps for team sizes $N=\{1,3,5\}$ (left to right).
In the $N=5$ case, the final blob appears only partially covered because the
supervisor terminated the run immediately after all four hotspots were detected,
before additional samples could accumulate around that region.
    }
    \label{fig:blob_coverage_team_size}
\end{figure}

\paragraph{Qualitative behaviour}
Visit heatmaps (Fig.~\ref{fig:visit_heatmap_team_size}) show increased exploration range with larger teams, but
also more concentrated revisits around hotspots at $N=5$. 
Figure~\ref{fig:blob_coverage_team_size} shows the corresponding blob-vs-coverage masks.
As expected, increasing the number of robots improves exploration efficiency and
hotspot discovery. In the five-robot run, the supervisor’s \texttt{STOP\_ALL}
condition (all four blobs detected) fired at $t=356$\,s. By contrast, the
$N=1$ and $N=3$ runs never reached this condition within the same simulation
horizon, leaving one or more blobs undiscovered. Larger teams clearly reduce
time-to-detection.

\paragraph{Quantitative metrics}
Quantitatively, Table~\ref{tab:team_size_metrics} shows that increasing team
size from $N=1$ to $N=5$ increases world coverage from $8.5\%$ to $18.1\%$ and
blob recall from $20.0\%$ to $65.6\%$. Precision improves sharply from
$N=1$ to $N=3$, but then drops slightly at $N=5$: the larger team discovers
more blob cells overall, but also generates more samples in background regions
around them. F1 follows a similar pattern: $N=3$ offers a good balance of
coverage and focus, while $N=5$ trades a small drop in F1 for full blob
detection and higher absolute recall.

\begin{table}[t]
\centering
\caption{Coverage and hotspot metrics for different team sizes}
\label{tab:team_size_metrics}
\begin{tabular}{lccc}
\hline
\textbf{Metric} & \textbf{$N=1$} & \textbf{$N=3$} & \textbf{$N=5$} \\
\hline
World coverage (\%)          & 8.5   & 9.8   & 18.1 \\
Blob recall (\%)             & 20.0  & 46.7  & 65.6 \\
Precision on coverage (\%)   & 17.0  & 34.4  & 26.1 \\
F1 score                     & 0.184 & 0.396 & 0.373 \\
Blobs detected ($\geq 1$ cell)     & 1/4   & 3/4   & 4/4 \\
Time to STOP\_ALL (s)        & --    & --    & 356 \\
\hline
\end{tabular}

\vspace{2pt}
\footnotesize
Time to STOP\_ALL is only defined when the supervisor's stopping condition
(all four blobs detected) is met. For $N=1$ and $N=3$ the runs reached the
simulation time limit without satisfying this condition; these entries are
therefore marked ``--''.
\end{table}

\paragraph{Failure modes}
For $N=1$ and $N=3$, limited spatial reach and stochastic drift can prevent the
team from ever entering all blob regions within the fixed time horizon, so
\texttt{STOP\_ALL} is not triggered. At $N=5$, although all blobs are detected,
stronger local interference produces revisit bands around blob boundaries where
multiple robots oscillate through the same high-richness region.

\subsection{Localisation Fidelity: PF Behaviour and PF--On vs PF--Off}
\label{sec:pf_results}

\begin{figure}[t]
    \centering
    \includegraphics[width=0.5\linewidth]{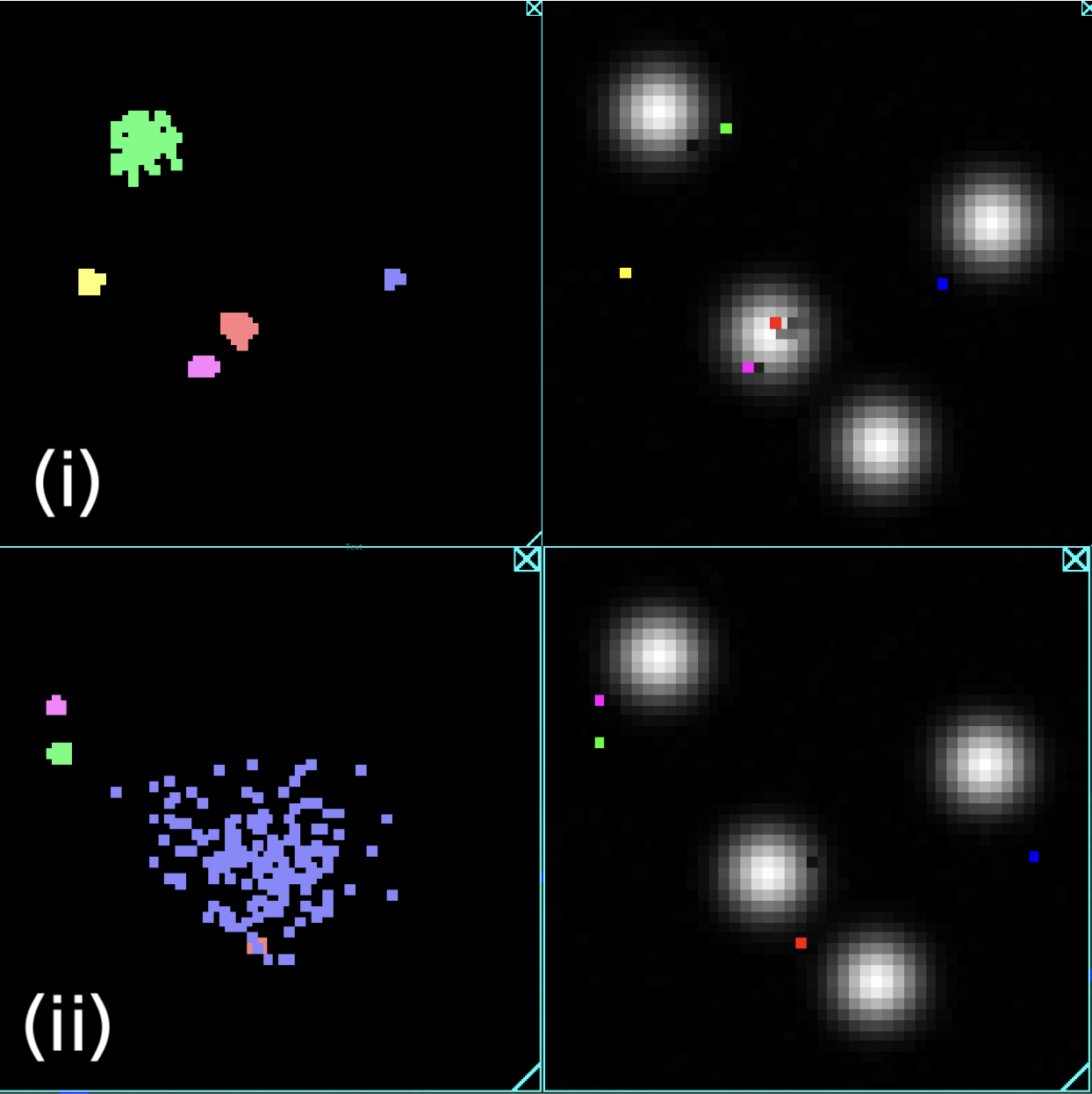}
    \caption{
Illustrative particle-filter behaviour during a five-robot run.
The left panel shows live particle clouds; the right panel shows true robot
poses overlaid on the richness map. Stable filters maintain tight clouds around
true poses; unstable filters spread and drift, producing large localisation error.}
    \label{fig:pf_behaviour}
\end{figure}

\paragraph{Qualitative behaviour}
Figure~\ref{fig:pf_behaviour} illustrates two characteristic PF outcomes during a five-robot run.
In the healthy case, the particle cloud expands under motion and collapses when GPS updates
agree with odometry, so robots maintain tight pose estimates and repeatedly enter BLOB mode
over hotspots. In the unstable case (mis-tuned resampling), the cloud spreads but never
collapses: weights remain almost flat, drift grows to metre scale, and robots start making
visibly wrong decisions (e.g.\ incorrect avoidance or missing hotspots). In the PF-disabled
ablation, \texttt{control\_x} and \texttt{SAMPLE\_AT} remain \texttt{(nan,nan)} and robots stay in
EXPLORE, traversing blob regions without reacting; visit-count maps are essentially empty.

\paragraph{Quantitative metrics}
Using the error metric in Eq.~\eqref{eq:loc_error}, a successful five-robot run
(all hotspots found) with $\alpha = 0.7$ achieved:
\[
\textbf{MAE = 0.133\,m},\quad
\textbf{RMSE = 0.208\,m},\quad
\textbf{Max = 1.105\,m}.
\]
An unstable run with a drifting PF yielded:
\[
\textbf{MAE = 0.363\,m},\quad
\textbf{RMSE = 0.914\,m},\quad
\textbf{Max = 3.783\,m},
\]
showing how a single diverging filter can dominate team-level localisation
error and distort behaviour. In the PF-off ablation, errors are undefined and no
valid samples are ever logged, which matches the visually empty heatmaps.

\paragraph{Failure modes}
These results confirm that RANT’s mapping pipeline is structurally dependent on
a coherent PF estimate: with mis-tuned resampling or degraded GPS the particle
cloud drifts, samples never register, and hotspots remain effectively invisible,
even though robots continue to move.

\subsection{Impact of Pheromone-Based Coordination}

\paragraph{Qualitative behaviour}
The virtual pheromone grid and blocked-mask coordination were evaluated by
comparing runs with coordination enabled versus disabled in a five-robot team.
Without pheromone guidance, robots frequently revisit already sampled regions,
especially around strong hotspots. This raises redundancy and slows the
discovery of new informative areas. With coordination enabled, robots show a
clear tendency to disperse: evaporation causes old paths to fade, the goal
selector penalises recently explored regions, and blocked discs prevent
repeated exploitation of confirmed hotspots.

\paragraph{Quantitative metrics}
Table~\ref{tab:coordination_comparison} quantifies these effects. Coordination
nearly doubles blob recall (33.9\% $\rightarrow$ 65.6\%) and increases world
coverage and F1 score while maintaining full hotspot detection. Although
absolute precision remains modest because robots must still survey large
background regions, coordination improves precision from $16.7\%$ to $26.1\%$.
Combined with the higher recall and F1, this indicates that blocked zones and
no-revisit logic make the swarm’s sampling more focused on blob cells rather
than simply amplifying random exploration.

\begin{table}[t]
\centering
\caption{Effect of Pheromone-Based Coordination on Mapping Quality (5 Robots)}
\label{tab:coordination_comparison}
\begin{tabular}{lcc}
\hline
\textbf{Metric} & \textbf{Coord.\ ON} & \textbf{Coord.\ OFF} \\
\hline
Blob cells (total)              & 180   & 180 \\
Blob cells covered              & 118   & 61 \\
Blob recall (\%)                & 65.6  & 33.9 \\
Covered cells (any)             & 452   & 366 \\
World coverage (\%)             & 18.1  & 14.6 \\
Precision on coverage (\%)      & 26.1  & 16.7 \\
Accuracy (whole world, \%)      & 84.2  & 83.0 \\
F1 score                        & 0.373 & 0.223 \\
Blobs detected ($\geq 1$ cell)        & 4/4   & 4/4 \\
\hline
\end{tabular}
\end{table}

\paragraph{Failure modes}
Even with coordination enabled, virtual pheromones do not eliminate all
redundant sampling: around very strong hotspots, robots can still cluster
briefly on blob boundaries before repulsion and evaporation pull them apart.
With coordination disabled, these effects are amplified, redundancy grows, and
the time to discover new blobs increases.

Overall, pheromone-based coordination improves coverage uniformity, reduces
redundant sampling, and makes better use of multi-robot deployments without
requiring explicit inter-robot communication.

\section{Conclusion }

RANT demonstrates that a small team of ant-inspired robots, equipped with
particle-filter localisation and simple virtual-pheromone rules, can reliably
discover and exploit hidden hotspots in a noisy, partially structured
environment. Larger teams accelerate hotspot detection but suffer diminishing
returns due to interference; localisation fidelity is structurally required for
mapping, and pheromone-based blocking improves coverage quality by reducing
redundant sampling and spreading robots across the arena.

The current GPS+IMU Webots proxy best reflects open or moderately occluded
sites; dense canopy would require additional visual–inertial or landmark cues
to keep drift bounded. The simulated world also simplifies vegetation,
dynamics, communication, and weather, and coordination is mediated by a central
supervisor rather than fully distributed communication.

Future work includes richer sensing (e.g.\ visual–inertial odometry, shared
local maps), more realistic communication and failure models, and field trials
on hardware platforms. This study is simulation-based and intended to isolate the effect of coordination and localisation assumptions under controlled conditions; transferring to physical robots will require additional modelling of sensing and communication constraints. Exploring alternative bio-inspired coordination
mechanisms or learning-based controllers would help test how far simple
ant-like rules can be pushed in large-scale, real-world monitoring tasks.

\vfill

\end{document}